\pdfoutput=1

\documentclass[11pt]{article}

\usepackage{acl}

\usepackage{times}
\usepackage{latexsym}
\usepackage{subcaption}
\usepackage{xcolor,colortbl}
\definecolor{SoftGreen}{HTML}{39ac73}

\usepackage[T1]{fontenc}

\usepackage[utf8]{inputenc}

\usepackage{microtype}

\usepackage{inconsolata}
\usepackage{tabularx}
\usepackage{booktabs}       
\usepackage[weather]{ifsym}
\usepackage{graphicx}
\usepackage[nameinlink,noabbrev,capitalise,sort]{cleveref}
\usepackage{todonotes}
\usepackage{fontawesome}

\usepackage{tikz}
\usepackage{bbding} 
\usepackage[weather]{ifsym}

\title{\noindent\parbox[][][t]{.02\linewidth}{\begin{tikzpicture}[color=white]
 at (0,0) {\includegraphics[width=.05\textwidth]{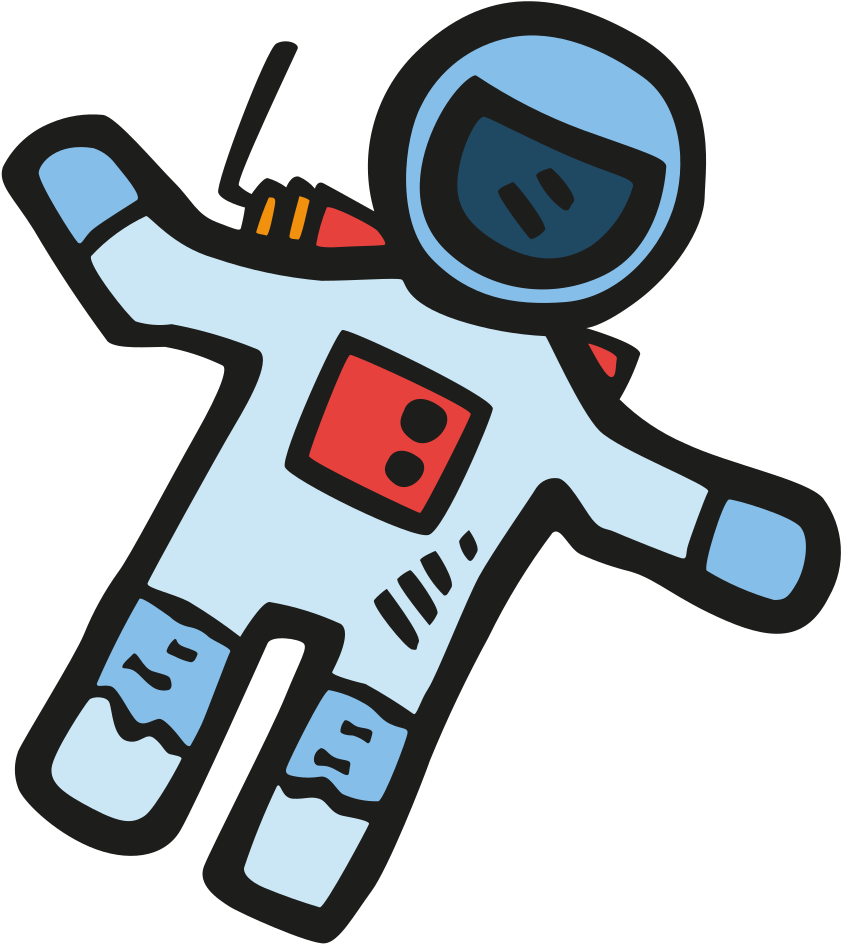}}
\end{tikzpicture}} Lost in Space: Probing Fine-grained Spatial Understanding in Vision and Language Resamplers}

\author{
    Georgios Pantazopoulos$^{1,2}$
    {\bf Alessandro Suglia$^{1,2}$}
    {\bf Oliver Lemon$^{1,2}$}
    {\bf \enspace Arash Eshghi$^{1,2}$}
    \AND \textnormal{$^1$Heriot-Watt University; $^2$Alana AI}\\
    \AND \textnormal {\texttt{\{gmp2000, a.suglia, o.lemon, a.eshghi\}}\texttt{@hw.ac.uk}}
}

\begin{document}
\maketitle
\begin{abstract}
An effective method for combining frozen large language models (LLM) and visual encoders involves a \textit{resampler} module that creates a `visual prompt' which is provided to the LLM, along with the textual prompt. 
While this approach has enabled impressive performance across many coarse-grained tasks like image captioning and visual question answering, \citep{alayrac2022flamingo, Dai2023InstructBLIPTG}, 
more fine-grained tasks that require spatial understanding have not been thoroughly examined.
In this paper, we use \textit{diagnostic classifiers} to measure the extent to which the visual prompt produced by the resampler encodes spatial information. 
Our results show that this information is largely absent from the resampler output when kept frozen during training of the classifiers. 
However, when the resampler and classifier are trained jointly, we observe a significant performance boost. 
This shows that the compression achieved by the resamplers can in principle encode the requisite spatial information, but that more object-aware objectives are needed at the pretraining stage to facilitate this capability\footnote{Code available \href{https://github.com/gpantaz/probing-resamplers}{here}}.
\end{abstract}

\section{Introduction}

Recent approaches for developing Vision and Language (V\&L) models leverage existing vision \citep{radford2021learning, fang2023eva, fang2023eva2}, and language experts \citep{touvron2023llama, zhang2022opt, touvron2023llama2} and try to learn a mapping between them \citep{alayrac2022flamingo, li2023blip, Dai2023InstructBLIPTG, you2023ferret, liu2023llava, liu2023improved}.
In most cases, the experts are kept frozen while the only learnable component is the mapping between the visual and the language expert.

The simplest approach uses a linear projection layer that matches the dimensionality of the visual and textual embeddings before feeding them to the LLM \citep{liu2023llava, liu2023improved}.
A more sophisticated method is to use a \textit{resampler} to compress the visual embeddings into a compact `visual prompt' that is then fed to the LLM either at the input level along with the text prompt \citep{li2023blip, Dai2023InstructBLIPTG} or via cross attention layers \citep{alayrac2022flamingo, li2023otter}.
From a practical standpoint, the resampler may accelerate training and inference as it significantly reduces the sequence length, but also facilitates in-context learning capabilities since additional examples can fit into the context window of the LLM.
As a result, these approaches have demonstrated impressive performance across multiple `coarse-grained' tasks such as image captioning, and visual question answering.

\begin{figure}[tb]
    \centering
    \includegraphics[width=\linewidth]{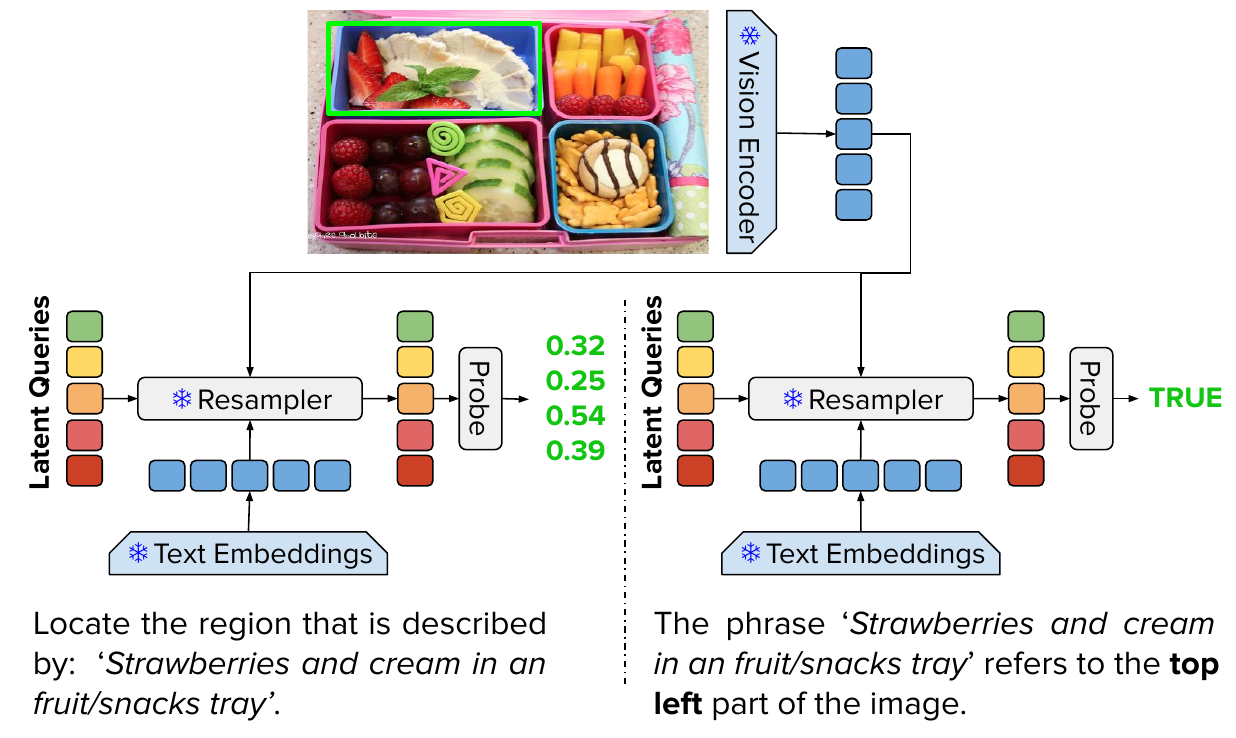}
    \caption{\textit{Explicit} (left) and \textit{implicit} (right) probing for spatial understanding. In the explicit setting, we probe for region localization, while in the implicit setting, the probe is trained to classify whether a description involving an image region is true of the image.}
    \label{fig:resampler}
\end{figure}

However, fine-grained tasks such as visual grounding and spatial understanding are relatively underexplored. Resamplers are usually pretrained on pairs of image-text data using contrastive learning \citep{li2023blip, Dai2023InstructBLIPTG}, and/or multimodal masked language modeling \citep{laurenccon2023obelisc, alayrac2022flamingo}, without relying on object-aware objectives. Given the importance of resamplers for the development of V\&L models, we ask whether this compression preserves fine-grained spatial information. Do the contrastive and language modeling objectives retain the overall scene structure, or is this information lost due to the absence of object-aware pretraining objectives? 


To address these questions, we train diagnostic classifiers to probe two different resampler modules for \textit{explicit} and \textit{implicit} spatial understanding — see ~\cref{fig:resampler}. 
Our results indicate that the multimodal resamplers do not facilitate spatial understanding. 
Nevertheless, in all settings, jointly fine-tuning the diagnostic classifiers and the resamplers significantly boosts performance, demonstrating that the compression achieved by the resamplers can in principle encode the requisite spatial information, but that more object-aware pretraining objectives are needed to facilitate this.

\section{Related Work}
\label{sec:related-work}

\paragraph{Resamplers} The idea of the resampler is inspired primarily by computer vision, where an attention mechanism is used to compress visual features into learnable queries (often referred to as slots) \citep{carion2020end, kamath2021mdetr, locatello2020object}. More recently, resamplers have been applied to more multimodal tasks. 
Flamingo \citep{alayrac2022flamingo} and subsequent open-source variants \citep{laurenccon2023obelisc, li2023otter} are based on the Perceiver Resampler \citep{jaegle2022perceiver}, with cross-attention between the latent queries and the visual embeddings followed by a stack of self-attention blocks that operate on the latent queries. In the Q-Former \cite{li2023blip, Dai2023InstructBLIPTG}, the latent queries are also informed by the input text and, therefore, create a more `linguistically informed' visual prompt.

\paragraph{Probing} Probing is a class of methods for interpreting neural models by assessing whether the model representations encode specific kinds of information at different processing stages \citep{belinkov2022probing}. 
The concept of probing is straightforward; we extract representations from a model that is already trained on some task(s), and use a lightweight \textit{diagnostic classifier} on top of these representations to solve a probing task that reflects the information that we seek to find.
The classifier’s performance is then taken to correlate with the extent to which that information is encoded by the model \citep{conneau2018you, hupkes2018visualisation}. 
Many within (multimodal) NLP have thus adopted probing to interpret model behavior \cite{kajic2022probing, salin2022vision,lindstrom2020probing}.


\begin{table*}[tb]
    \centering
    \begin{tabular}{@{}l cc cc cc@{}}
    \toprule
         & \multicolumn{2}{c}{RefCOCOg} & \multicolumn{2}{c}{VSR random} & \multicolumn{2}{c}{RCM} \\
         \cmidrule(lr){2-3}
         \cmidrule(l){4-5}
         \cmidrule(l){6-7}
         & Validation & Test & Validation & Test & Validation & Test\\
    \midrule
    Random & - & -  & - & 50.00 & 50.00 & 50.00 \\
    Human & - & -  & - & 95.40  & - & 92.29\\
    \midrule
    MDETR \citep{kamath2021mdetr} & 83.35 & 83.31 & - & - & -  & -\\
    CLIP$^*$ \citep{radford2021learning} & - & - &  - & 56.0 & -  & -\\
    Unitab \citep{yang2022unitab} & 84.58 & 84.70 & - & - & - & - \\
    ViLT \citep{kim2021vilt} & 69.14 & 68.93 & 71.38 & 71.53 & 83.16  & 83.25 \\ 
    \midrule
       \textcolor{blue}{\SnowflakeChevron} Q-Former & 30.39 & 30.26 & 66.91 & 64.97 & 70.12 & 69.49\\ 
       \textcolor{red}{\faFire} Q-Former & 71.47 & 71.72 & 80.86 & 80.50 & 81.68 & 81.35\\
       \textcolor{blue}{\SnowflakeChevron} IBLIP Q-Former & 20.00 & 19.92 & 58.07 & 55.72 & 64.58 & 63.08\\
       \textcolor{red}{\faFire} IBLIP Q-Former & 68.89 & 69.34 & 78.40 & 76.99 & 83.11 & 80.86\\
     \bottomrule
    \end{tabular}
    \caption{Linear probing results. \textcolor{blue}{\SnowflakeChevron}/\textcolor{red}{\faFire} denotes that the resampler is frozen/unfrozen. $^*$ results from \citet{liu2023visual}.}
    \label{tab:linear-prob-results}
\end{table*}

\begin{figure*}
    \centering
    \begin{subfigure}[b]{.49\textwidth}
        \includegraphics[width=1\linewidth]{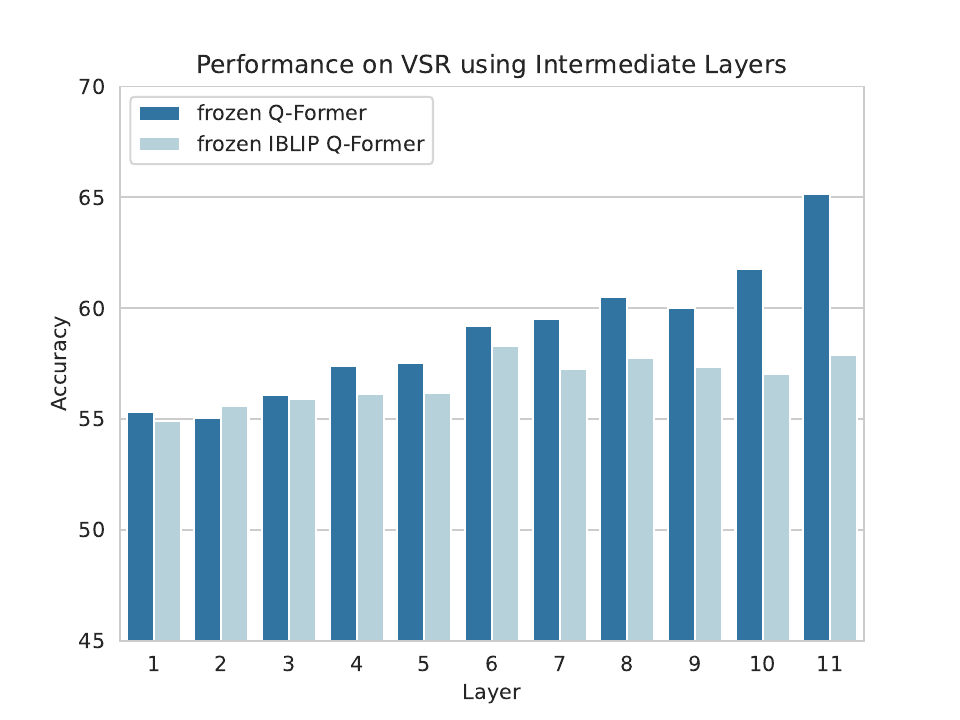}
        \caption{}
        \label{fig:interm-layers}
    \end{subfigure}%
    \hfill
    \begin{subfigure}[b]{.49\textwidth}
        \includegraphics[width=1\linewidth]{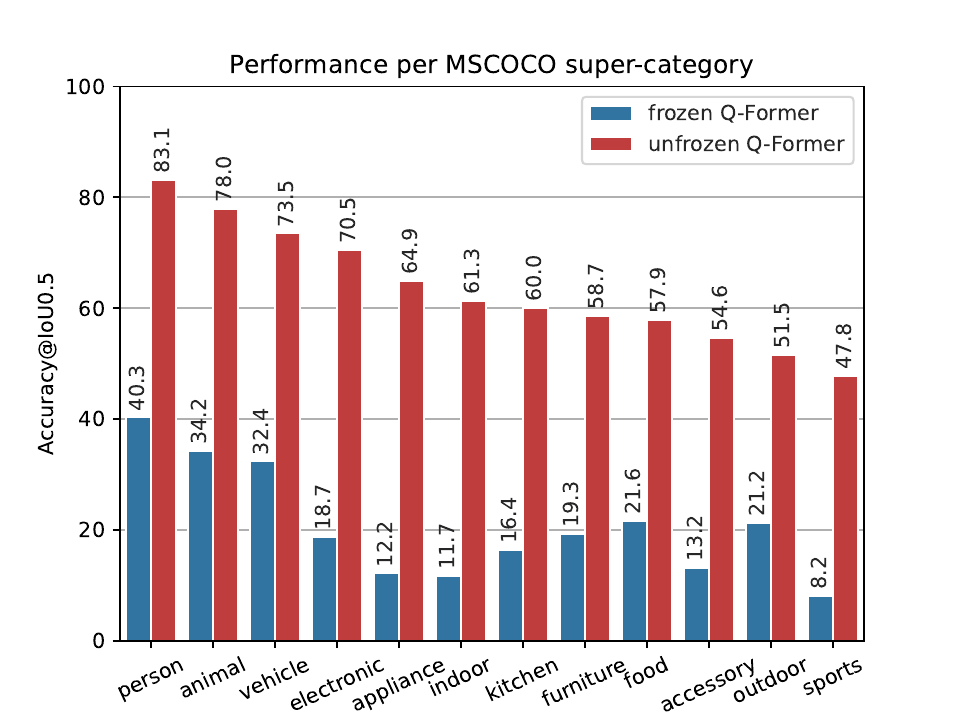}
        \caption{}
        \label{fig:refcocog-category-results}
    \end{subfigure}
    \caption{Performance on (a) VSR per intermediate layer, (b) RefCOCOg per MSCOCO super-category.}
\end{figure*}


\section{Experiments}
\paragraph{Is spatial understanding a property of V\&L resamplers?}
We experiment with three different spatial understanding tasks. In RefCOCOg \citep{mao2016generation}, the objective is to predict the coordinates of the region that is described by the input phrase. Secondly, we use the `random split' from the VSR dataset \citep{liu2023visual}, where the model has to assess the validity of a caption describing a spatial relationship between two entities.
Finally, we introduce the Region Cell Matching (RCM) task, which follows the VSR formulation but is designed to test for a more rudimentary form of spatial understanding regarding the location of one entity in the image.
Inspired by CAPTCHAs, an image is divided into a 3x3 grid, and each grid cell is assigned a location description (such as top left or middle).  
We generate synthetic captions by combining RefCOCOg descriptions with the cell location as shown in the implicit probing example of \cref{fig:resampler}.
To ensure that performance is not influenced by frequency biases, we balanced the distribution of positive and negative examples. 
\cref{appendix:grid-cell} contains further details about the dataset.

In our experiments, we use the Q-Former from the first pretraining stage of BLIP2 \citep{li2023blip} and InstructBLIP \citep{Dai2023InstructBLIPTG}. 
To probe the resamplers, we follow past work \cite{belinkov2022probing} and use a single linear layer after flattening the embeddings of the query tokens. 
For RefCOCOg, the linear layer predicts the normalized coordinates of the region that matches the referring expression. 
We use the bounding box loss from (M)DETR \cite{carion2020end, kamath2021mdetr}: a weighted sum of the Generalised IoU and L1 losses. 
Similarly, for VSR and the RCM task, we use a linear layer that predicts the probability that the query matches the image trained using binary cross entropy.
We tune the learning rate, number of epochs, and loss weights (only for RefCOCOg) using Bayesian hyperparameter optimization \cite{pmlr-v28-bergstra13} for at least ten iterations. For further implementation details, see \cref{appendix-impl}.
In all cases, we evaluate the best model in terms of validation performance.


We compare the two resamplers against similarly-sized models that employ patch representations. We avoid comparison against models with object-centric visual encoding because the task of visual grounding is significantly easier in these models as they need to select the correct candidate bounding box provided from the detector as opposed to explicit image region prediction. Additionally, we provide results where the linear classifier is jointly trained along with the resampler as an upper bound for the performance with frozen representations.


\begin{table*}[tb]
    \centering
    \small
    \begin{tabular}{@{}lccccccc@{}}
    \toprule
    Category & Adjacency & Directional & Orientation & Projective & Proximity & Topological & Unallocated \\
    \midrule
    \textcolor{blue}{\SnowflakeChevron} Q-Former &  \cellcolor{SoftGreen!45} 61.94 &  \cellcolor{SoftGreen!17} 42.05 &  \cellcolor{SoftGreen!38} 56.93 &  \cellcolor{SoftGreen!46} 62.87 &  \cellcolor{SoftGreen!43} 60.15 &  \cellcolor{SoftGreen!63} 74.56 &  \cellcolor{SoftGreen!54} 68.42 \\
    \textcolor{red}{\faFire} Q-Former &  \cellcolor{SoftGreen!55} 68.86 &  \cellcolor{SoftGreen!64} 75.00 &  \cellcolor{SoftGreen!53} 67.15 &  \cellcolor{SoftGreen!68} 78.29 &  \cellcolor{SoftGreen!74} 81.95 &  \cellcolor{SoftGreen!77} 83.94 &  \cellcolor{SoftGreen!60} 72.37 \\
    \textcolor{blue}{\SnowflakeChevron} IBLIP Q-Former &  \cellcolor{SoftGreen!39} 57.44 &  \cellcolor{SoftGreen!12} 38.64 &  \cellcolor{SoftGreen!40} 58.39 &  \cellcolor{SoftGreen!34} 54.21 &  \cellcolor{SoftGreen!15} 40.60 &  \cellcolor{SoftGreen!51} 66.14 &  \cellcolor{SoftGreen!32} 52.63 \\
    \textcolor{red}{\faFire} IBLIP Q-Former &  \cellcolor{SoftGreen!47} 62.98 &  \cellcolor{SoftGreen!54} 68.18 &  \cellcolor{SoftGreen!54} 67.88 &  \cellcolor{SoftGreen!63} 74.61 &  \cellcolor{SoftGreen!69} 78.95 &  \cellcolor{SoftGreen!75} 83.15 &  \cellcolor{SoftGreen!68} 77.63 \\
    \bottomrule
    \end{tabular}
    \caption{VSR results per model for different categories of spatial relationships. \textcolor{blue}{\SnowflakeChevron}/\textcolor{red}{\faFire} denotes that the resampler is frozen/unfrozen.}
    \label{tab:vsr-category-results}\vspace{-0.3cm}
\end{table*}

\cref{tab:linear-prob-results} shows the results for the models that we are considering.
We observe that both resamplers perform poorly on RefCOCOg  when kept frozen, and, therefore, are unable to perform explicit visual grounding. 
A possible counter-argument could be that predicting raw coordinates within the image is too difficult to solve with a single linear layer.
However, we observe similar trends with VSR and RCM, which test for spatial understanding in an easier binary classification setup.
While the resamplers perform better than random baselines in these tasks, there is a significant gap between the performance of the frozen and fine-tuned backbones.
We believe this is an outcome of the pretraining objectives of the Q-Former that do not explicitly facilitate fine-grained object-centric representations. This is in line with previous work, which found that V\&L models trained with contrastive objectives act as bag-of-words and do not preserve spatial information~\cite{yuksekgonul2022and}.
On the other hand, the significant boost achieved by unfreezing the resamplers shows that the compression of the input embeddings is, in principle, able to capture spatial information and, therefore, that the resampler as an architectural choice does not necessarily constitute a bottleneck. 

\paragraph{Is spatial information encoded in earlier layers but discarded in deeper layers?} We previously observed that resamplers have poor performance in spatial understanding tasks when using representations from the last layer. 
Next, we examine if the representations from intermediate layers better encode spatial information.
Intuitively, representations from earlier layers could lead to greater probing performance as they are closer to the visual encoder's output.
\cref{fig:interm-layers} shows the results on VSR after probing representations from intermediate layers.
Overall, intermediate layer representations do not provide performance gains. There is a clear upward trend regarding the performance of the Q-Former from BLIP2, whereas for InstructBLIP we observe fluctuations within a small range across layers. A similar trend is observed in the RefCOCOg results which are included in \cref{sec:appendix}.

\paragraph{Scaling the Probing Classifier}
Additionally, we experiment with scaling the probe classifier by introducing non-linearities. In particular, we use 2-layer and 4-layer classifiers with SwiGLU activation functions. We refrain from using more complex classifiers because they may infer features that are not actually used by the underlying model \citep{hupkes2018visualisation}. For training, we used the same setup as with our previous experiments.

\cref{tab:scaling} illustrates the results with increasing prompt complexity. While we observe a common trend of increasing performance when we make the probe more complex, the accuracy of the non-linear probes does not indicate that the resampler encodes spatial information which can be easily retrieved. Additionally, the performance gap between the simplest and the most complex probe in the case of InstructBLIP indicates that fine-grained spatial understanding is `built-up' within the probe and is not necessarily a property of the resampler component.

\begin{table}[tb]
    \centering
    \scriptsize
    \begin{tabular}{@{}lc cc cc@{}}
    \toprule
         Model & \#Layers & RefCOCOg & VSR random & RCM \\
    \midrule
       \textcolor{blue}{\SnowflakeChevron} Q-Former \\ 
       & 1 & 30.26 & 64.97 & 69.49\\
       & 2 & 32.08 & 65.15 & 69.98\\
       & 4 & 34.49 & 65.01 & 70.71\\
    \midrule
       \textcolor{blue}{\SnowflakeChevron} IBLIP Q-Former \\ 
       & 1 & 19.92 & 55.72 & 63.08\\
       & 2 & 25.01 & 58.09 & 68.66\\
       & 4 & 34.49 & 59.09 & 69.29\\
     \bottomrule
    \end{tabular}
    \caption{Probing results by scaling the probing classifier.}
    \label{tab:scaling}
\end{table}

\subsection{Discussion}
\paragraph{Performance analysis per object category} \cref{fig:refcocog-category-results} illustrates the Q-Former's performance on RefCOCOg per MSCOCO \citep{lin2014microsoft} super-category. We observe that the frozen/unfrozen resamplers behave differently but also have significant variation between object categories. To further understand the possible reasons for this variation, we computed the Kendall coefficient \citep{kendall1938new} between the performance of each super-category and 1) the distribution of train examples, 2) the area of each bounding box, 3) and the distance of the bounding box from the center of the image (\cref{tab:refcocog-correlations}). Interestingly, the main factor that correlates positively with the performance per category is the area of the bounding box. We also observe that 
the further the bounding box deviates from the center, the more the performance drops. These two observations imply that the Q-Former constructs the visual prompt by `summarizing' the most central entities within an image, ignoring positional outliers.

\paragraph{Which spatial relationships are difficult to capture?} In \cref{tab:vsr-category-results}, we break down the VSR results according to the spatial relationship type. Both resamplers perform the best in topological relations across frozen/unfrozen conditions. Directional relations seem challenging for out-of-the-box resamplers, though this relation can be captured during fine-tuning. Finally, captions describing adjacency or orientation properties are difficult even for fine-tuned resamplers.

\paragraph{Effect of learning objectives} We showed that multimodal resamplers pretrained with contrastive learning and multimodal language modeling objectives do not capture spatial information well. These are undoubtedly important objectives as they enable large-scale pretraining, however, on their own, they are not sufficient for enabling fine-grained spatial understanding.

Finally, we observed that BLIP-2's Q-Former consistently outperformed the one from InstructBLIP. 
However, as shown in \cref{fig:interm-layers}, the performance of the two resamplers is comparable for early layers.
We hypothesize that during instruction tuning, the InstructBLIP Q-former may get away with providing even less fine-grained information since the language modeling loss is already low due to the high-quality LLM, leading to a forgetting effect \citep{mccloskey1989catastrophic}.

\section{Conclusion}
In this paper, we explored to what degree multimodal resamplers preserve spatial information. 
While previous work has demonstrated the effectiveness of resamplers across a variety of V\&L tasks, our investigation revealed  their limitations when applied to spatial understanding tasks.
In particular, we probed two resamplers and showcased that grounding natural language descriptions in image regions is not an inherent ability of these modules.
Furthermore, probing experiments showed limited spatial understanding in two easier settings. These involved image-text matching with captions referencing the absolute location of an entity, or spatial relationships between two entities.
Nevertheless, our results showcased that when the resampler is fine-tuned, the compression of the visual encoding induced by the resampler can be effective.
We believe that this is due to the lack of an object-aware pretraining objective that would encourage the resamplers to encode spatial information. 
Future work should build upon our findings and design objectives that incentivize disentangled representations \cite{bengio2013representation}.

\section*{Limitations}





This study centered on exploring some architectural components of current V\&L models with regard to their ability to encode spatial information.
For the purpose of our study, it is necessary that the visual and textual representations are already fused. 
Models adopting unimodal resamplers do not facilitate this because 1) the fusion happens only in the successive cross-attention layers of the LLM \citep{alayrac2022flamingo}, 
or 2) the visual embeddings are concatenated with the text embeddings at the input of the LLM \citep{bai2023qwen}. 
While we could extract representations from intermediate layers from a model like IDEFICS \citep{laurenccon2023obelisc}, this would have been an unfair comparison with BLIP-2 style  models because the former adds more layers to the original resampler architecture. 
The other option would be to provide the visual embeddings and the text embeddings to the probe, but this defeats the purpose of the probing classifier as probe since it would have to perform the necessary multimodal fusion internally; thus making any comparisons uninterpretable.
Consequently, our study does not encompass the entirety of available models adopting resamplers, and the findings may not be fully representative of the broader V\&L model landscape.

We also recognize the limitation in our exploration of spatial understanding as an emergent ability in V\&L models.
The question of whether spatial understanding materializes as a natural consequence of model scale remains unanswered in our study.
A more in-depth investigation controlling the pretraining dataset, the size of the models as well, and the training hyperparameters is required in order to truly understand the capacity of these models to develop fine-grained and disentangled representations that facilitate spatial understanding.

\section*{Acknowledgements}
We would like to thank the reviewers for their valuable feedback during the ARR process. Additionally, we would like to thank Malvina Nikandrou, and Ioannis Konstas for their suggestions with regards to the experimental setup.
This work was supported by the Edinburgh International Data Facility (EIDF) and the Data-Driven Innovation Programme at the University of Edinburgh.

\bibliography{acl}
\newpage

\appendix
\section{Region Cell Matching}\label{appendix:grid-cell}
\begin{figure}[ht!]
    \centering
    \begin{subfigure}[b]{.49\textwidth}
        \includegraphics[width=1\linewidth]{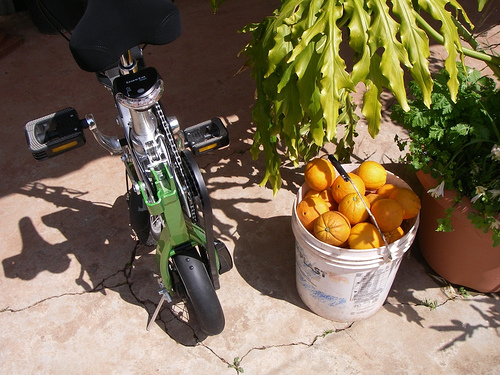}
        \caption{The phrase `A earth tone flower pot with a green bush in it.' refers to the \textbf{middle right} part of the image.}
        \label{fig:grid-cell-pos}
    \end{subfigure}%
    \hfill
    \begin{subfigure}[b]{.49\textwidth}
        \includegraphics[width=1\linewidth]{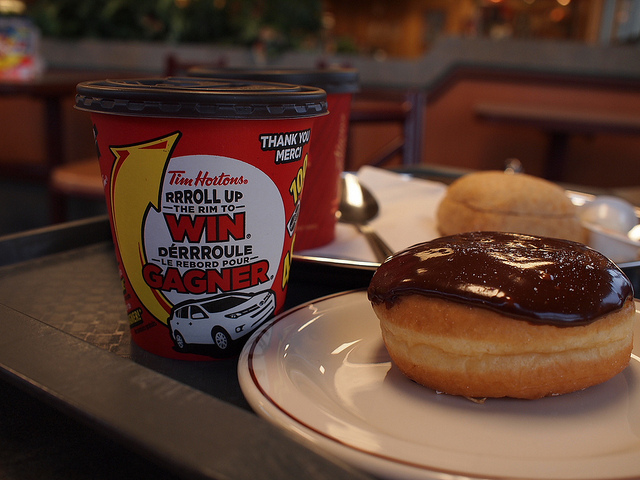}
        \caption{The phrase `A tan and brown donut with a thick coating of chocolate on top.' refers to the \textbf{middle} part of the image.}
        \label{fig:grid-cell-neg}
    \end{subfigure}
    \caption{Illustration of positive (a) and negative (b) examples from the RCM task.}
    \label{fig:grid-cell-examples}
\end{figure}

The purpose of Region Cell Matching (RCM) is to evaluate a model's capacity to perform visual grounding in an implicit manner, i.e., the model does not need to provide a specific region within an image, but it has to substantiate if a given description belongs to a certain region within an image. To make the task even easier, this region is not arbitrary, it corresponds to one of the cells of a $3\times3$ grid on top of the image. Each of these cells is mapped to a natural language description, for example, top left, middle, top right, etc. \cref{fig:grid-cell-examples} illustrates one positive and one negative example from the dataset.

To create the dataset, we started from RefCOCOg examples and assigned each bounding box to one of the cells within the grid by matching its center to the closest cell. 
To prevent overpopulating the dataset with examples where the bounding box is centered, we downsampled the dataset so that the distribution of the cells is balanced. With this process, we created a subset of $N$ positive examples that are evenly distributed between the 9 cells. 
In order to prevent biases related to the distribution of the cells we additionally created $N$ negative examples as follows: For each grid cell $i$ with $N_i$ positive examples we selected $N_i / 8$ from every other cell $j$ as negative examples.
We repeat the steps for train, validation, and test sets resulting in 46k, 3k, and 5.5k samples, respectively.

\paragraph{Human Performance} Apart from fine-tuning ViLT, we established a human baseline by estimating the performance of humans in the task.
We developed a Gradio interface \citep{abid2019gradio} where participants received the input image, the region description as well as the assigned cell and they were asked to provide a binary response to the question `Does the phrase match the location in the image?'. In order to imitate the training and evaluation setting in our experiments, we did not provide any additional information (e.g., there was no visible grid on top of the image as this would have trivialized the task) to the participants, with the exception of a few introductory examples before actually completing the task.

Since a region may overlap with multiple grid cells, we also gave participants the option to provide up to 4 grid cells ranked in terms of priority.
Additionally, participants may refrain from answering the question if the phrase is factually incorrect (e.g., the phrase `A dog with a frisbee' is factually incorrect if there is no dog within the image). We decided to include this option to avoid any potential confusion and introduce unnecessary noise to the annotation.

We recruited a total of five participants who were informed about the study and the use of their data. Each participant annotated $100$ examples from the test set ($50$ positive / $50$ negative).
To estimate a human baseline, we removed the instances where each annotator assigned either multiple cells or labeled an instance as factually incorrect.
Finally, we measured the annotator agreement with the Fleiss' kappa coefficient \citep{fleiss1971measuring}: $k = 80.98$.

\begin{table*}[tb]
    \centering
    \begin{tabular}{@{}llcccccc@{}}
    \toprule
    Task &  \multicolumn{2}{c}{Hyperparameters} & \multicolumn{2}{c}{Q-Former} & \multicolumn{2}{c}{IBLIP Q-Former}\\
         \cmidrule(l){2-3}
         \cmidrule(l){4-5}
         \cmidrule(l){6-7}
         & Name & Value & \textcolor{blue}{\SnowflakeChevron} & \textcolor{red}{\faFire} & \textcolor{blue}{\SnowflakeChevron} &  \textcolor{red}{\faFire}\\
    \midrule
    RefCOCOg & lr & [1e-5, 5e-4] & 4.85e-4 & 1.03e-4 & 2.55e-4 & 1.08e-4\\
             & epochs & \{20, 30, 40\} & 40 & 20 & 40 & 40\\
             & GIoU scale & \{1, 2x, $x \in \{1,\dots, 10\}$\} & 6 & 20 & 16 & 20\\
             & L1 scale & \{1, 2x, $x \in \{1,\dots, 10\}$\} & 20 & 18 & 18 & 8\\
    \midrule
    VSR & lr & [1e-5, 5e-4] & 3.92e-4 & 4.59e-4 & 1.03e-4 & 2.49e-5\\
        & epochs & \{3, 5, 10, 15, 20\} & 5 & 10 & 15 & 20\\
    \midrule
    RCM & lr & [1e-5, 5e-4] & 1.94e-5 & 4.74e-4 & 4.34e-4 & 3.10e-5\\
        & epochs &  \{50, 100, 150\} & 100 & 150 & 150 & 50\\
    \bottomrule
    \end{tabular}
    \caption{Hyperparameters used during bayesian optimization. Additionally, we performed early stopping for RCM with a patience of 10 epochs.}
    \label{tab:bayesian-opt}
\end{table*}

\section{Implementation Details}
\label{appendix-impl}
In our experiments we used BLIP2's \citep{li2023blip} Q-Former from the first pretraining stage which is pretrained using contrastive, image-text matching, and masked language modeling losses. In this stage the Q-Former is trained as a standalone component, i.e, there is no language modeling loss from an LLM.
For InstructBLIP \citep{Dai2023InstructBLIPTG}, we used the Q-Former that is trained to prompt the Vicuna-7B model \citep{vicuna}.

For all experiments we used AdamW optimizer with weight decay of 0.01 and 10\% warmup. We used a fixed batch size of $128$ and tuned exclusively the learning rate and the number of steps following \citep{tuningplaybookgithub}. For RefCOCOg we also tuned the scale of GIoU and L1 loss. \cref{tab:bayesian-opt} shows the hyperparameters that were tuned, their minimum and maximum values, and the best configuration for frozen and unfrozen resamplers. 
All training logs regarding the main experiments as well as the experiments using intermediate representations are available \href{https://api.wandb.ai/links/gpantaz/4amym475}{here}.

\section{Additional Results}
\label{sec:appendix}
\begin{figure}[tb]
    \centering
    \includegraphics[width=\linewidth]{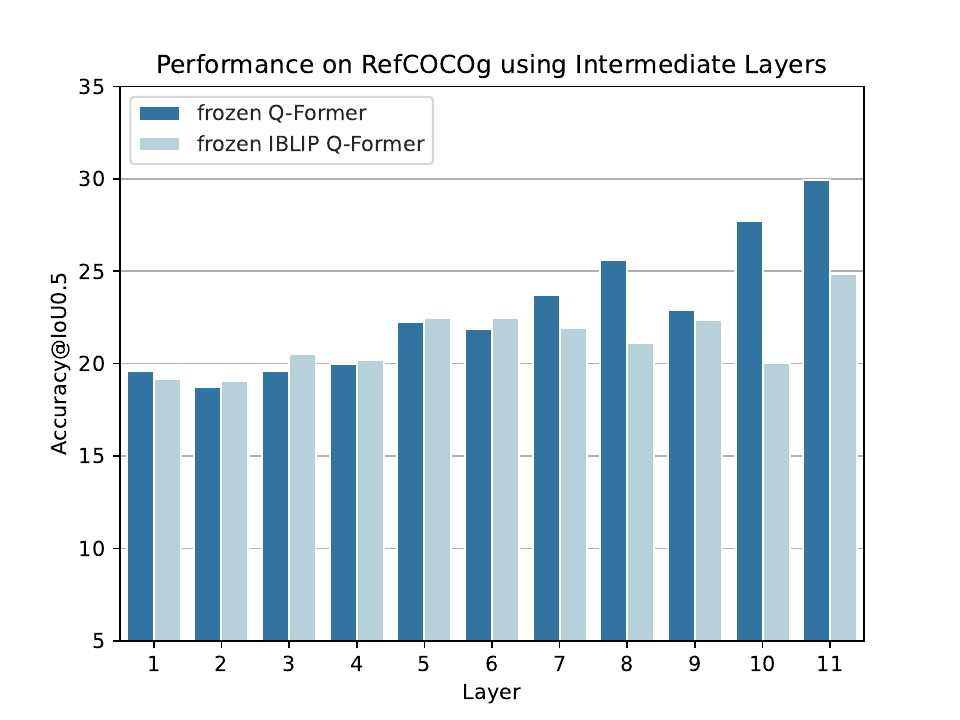}
    \caption{Performance of Q-Former on RefCOCOg per intermediate layer.}
    \label{fig:refcocog-layer-plot}
\end{figure}

\begin{figure}[tb]
    \centering
    \includegraphics[width=\linewidth]{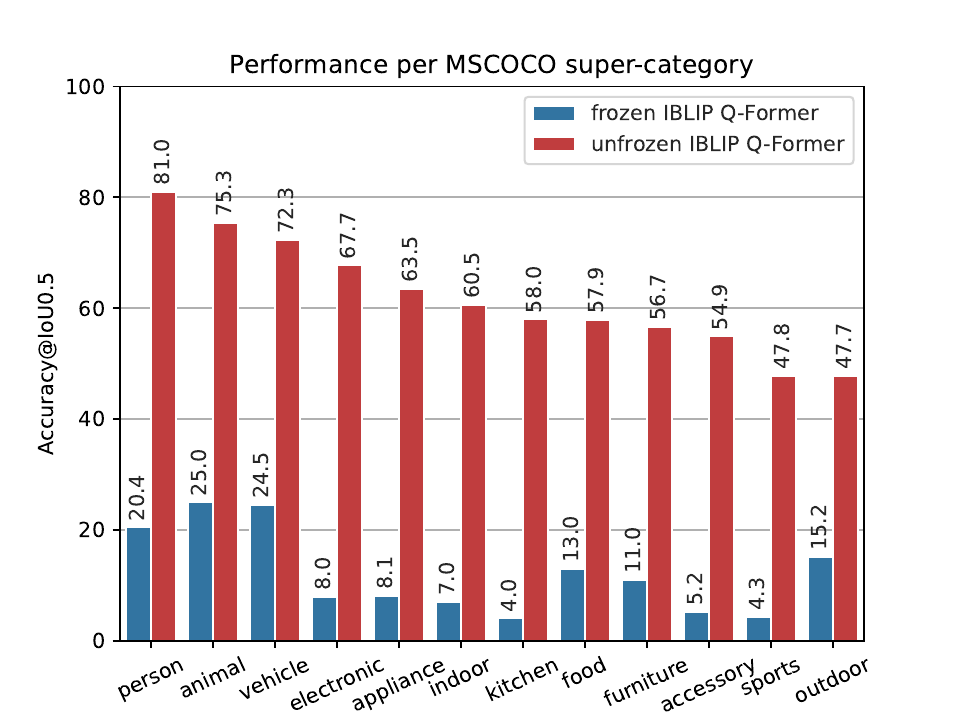}
    \caption{Performance of InstructBLIP Q-Former on RefCOCOg per MSCOCO super-category.}
    \label{fig:iblip-refcocog-super-category}
\end{figure}

\begin{table}[h!]
    \centering
    \small
    \begin{tabular}{@{}llcccc@{}}
    \toprule
     & \multicolumn{2}{c}{Q-Former} & \multicolumn{2}{c}{IBLIP Q-Former}\\
         \cmidrule(l){2-3}
         \cmidrule(l){4-5}
         & \textcolor{blue}{\SnowflakeChevron} & \textcolor{red}{\faFire} & \textcolor{blue}{\SnowflakeChevron} &  \textcolor{red}{\faFire}\\
    \midrule
    \# examples (train) & 0.63 & \textcolor{gray}{0.42} & \textcolor{gray}{0.33} & \textcolor{gray}{0.42}\\
    Area (test) & 0.84 & 0.45 & 0.66 & 0.45\\ 
    Distance (test) & -0.51 & \textcolor{gray}{-0.42} & -0.63 & -0.42\\
    \bottomrule
    \end{tabular}
    \caption{Kendall correlation coefficient between performance of resamplers and 1) \# training examples, 2) bounding box area of test examples, and 3) distance between the center of the bounding box and the center of an image of test examples. \textcolor{gray}{Numbers} illustrate p-values greater than 0.05.}
    \label{tab:refcocog-correlations}
\end{table}

For completeness, \cref{fig:refcocog-layer-plot} shows the results on RefCOCOg after obtaining the representations of the queries from intermediate layers. 
We observe a similar pattern as in \cref{fig:interm-layers}, where there is a clear boost when obtaining the representations from deeper layers from the BLIP2's Q-Former but in the case of the InstructBLIP we observe fluctuations in the performance.

\paragraph{RefCOCOg performance analysis per object category} In order to better understand variations in performance between the different object categories, we used the distribution of 1) \# training examples, 2) bounding box area of test examples, and 3) distance between the center of the bounding box and the center of an image of test examples. \cref{tab:refcocog-correlations} shows the Kendall correlation coefficient between the performance on different super-categories and the three conditions. 

\paragraph{Relationship between performance of probe and the visual LLM}
With regards to the relation between probing and the performance of the visual LLM, we prompted InstructBLIP on VSR and RCM with the prompts reported in the original paper and ranked the logits for positive / negative answers.
The performance of the InstructBLIP model is 61\% on VSR and 51\% on RCM.
While this is a performance increase in the case of probing on VSR, it shows that even the full stack of the MLMM is unable to robustly retrieve spatial information from the compressed visual sequence.
In the case of RCM we observe a notable drop which we assume is due to the lack of any similar tasks during the instruction-tuning phase.

\end{document}